\documentclass[10pt,twocolumn,letterpaper]{article}

\usepackage[pagenumbers]{cvpr} 

\usepackage{graphicx}
\usepackage{amsmath}
\usepackage{amssymb}
\usepackage{booktabs}
\usepackage{algorithm}
\usepackage{algpseudocode}
\usepackage{float}
\algrenewcommand\algorithmicrequire{\textbf{Input:}}
\algrenewcommand\algorithmicensure{\textbf{Output:}}

\usepackage[pagebackref,breaklinks,colorlinks]{hyperref}

\usepackage[capitalize]{cleveref}
\crefname{section}{Sec.}{Secs.}
\Crefname{section}{Section}{Sections}
\Crefname{table}{Table}{Tables}
\crefname{table}{Tab.}{Tabs.}

\def\cvprPaperID{*****} 
\def\confName{CVPR}
\def\confYear{2023}

\begin{document}

\title{Image Harmonization with Diffusion Model}

\author{Jiajie Li$^1$ \quad Jian Wang$^2$ \quad Chen Wang$^3$ \quad Jinjun Xiong$^1$ \\
$^1$University at Buffalo \quad $^2$Snap Inc. \quad $^3$IBM Research \\
}
\maketitle

\begin{abstract}
Image composition in image editing involves merging a foreground image with a background image to create a composite. Inconsistent lighting conditions between the foreground and background often result in unrealistic composites. Image harmonization addresses this challenge by adjusting illumination and color to achieve visually appealing and consistent outputs. In this paper, we present a novel approach for image harmonization by leveraging diffusion models. We conduct a comparative analysis of two conditional diffusion models, namely Classifier-Guidance and Classifier-Free. Our focus is on addressing the challenge of adjusting illumination and color in foreground images to create visually appealing outputs that seamlessly blend with the background. Through this research, we establish a solid groundwork for future investigations in the realm of diffusion model-based image harmonization.
\end{abstract}  

\label{sec:intro}
\section{Introduction}

Image composition, a common operation in image editing, involves merging a foreground image with a background image to create a composite. However, inconsistent lighting conditions between the foreground and background often result in unrealistic composites.

Image harmonization aims to adjust the appearance of the foreground image to achieve compatibility with the background, creating a natural and realistic composite. Inharmonious elements, such as differences in color, illumination, and texture, can violate natural laws and disrupt visual coherence. Achieving image harmonization without altering the structure or semantics of the composite image is a crucial and challenging task.

Traditional methods for image harmonization focused on color transformation to match the foreground's color statistics with the background. While efficient, these methods often fail to capture realism adequately. Recent deep learning approaches have utilized end-to-end image transformation to improve harmonization quality. These methods\cite{sunkavalli2010multiscale,tsai2017deep,cun2020improving}, such as encoder-decoder frameworks with U-Net\cite{ronneberger2015u} structures, capture semantic and low-level features for generating more realistic images. However, they are often limited to supervised learning settings.

In this work, we explore both classifier-guided\cite{dhariwal2021diffusion} and classifier-free\cite{ho2022classifier} conditional diffusion models for image harmonization tasks. By conditioning on unharmonized images, our image-to-image diffusion models generate high-quality outputs with realistic and consistent colors. Our classifier-free approach begins by training Denoising Diffusion Probabilistic Models (DDPM) \cite{ho2020denoising} in an end-to-end manner. We then employ Latent Diffusion Models (LDM)\cite{rombach2022high} that utilizes ControlNet\cite{zhang2023adding} for fine-tuning the pre-trained Stable Diffusion\cite{rombach2022high} model. The LDM model harmonizes the input image on the latent space, resulting in high-fidelity outputs. To overcome the issue of detail loss in the results obtained from Stable Diffusion, we combine the classifier-guidance method to ensure the appearance consistency. We propose a method to selectively transfer color information from the generated images, making it adaptable to other tasks. To enhance composite image harmonization, we integrate background "light" using a straightforward brightness prediction method. This ensures the consistency of appearance in the generated images.


Our contributions are as follows: (1) we develop the first image harmonization diffusion model frameworks using both DDPM and LDM; (2) we analyze and address the challenges of latent diffusion models in image editing tasks, proposing universal methods to maintain appearance consistency by leveraging the classifier-guidance method; (3) we present comprehensive experiments demonstrating the effectiveness of our diffusion model-based approach, achieving significantly superior performance compared to previous methods in image harmonization.
\section{Related Works}
  \subsection{Image Harmonization}
    Image harmonization aims to adjust the foreground to make it consistent with the background to produce a realistic composite image. Recent image harmonization methods can be categorized into traditional methods and deep learning-based methods.
    
    Traditional methods use color transformation techniques to match the color statistics between the foreground and background. These methods include color distribution matching \cite{pitie2005n, reinhard2001color, cohen2006color}, color histogram transformation \cite{xue2012understanding}, multi-scale statistics analysis \cite{sunkavalli2010multiscale}, color clustering \cite{lalonde2007using}. It's easier to see that the major difference between those methods is how the image of an image is represented (e.g. distribution, histogram, etc.). These methods are fast and simple but often fail to handle complex scenarios and produce artifacts since the realism of the image is usually not well reflected in those statistics.
 
    Deep learning methods are adopted to mitigate the issue that hand-crafted features such as color statistics used in previous machine learning methods cannot well reflect the realism of the image.
    \cite{zhu2015learning} first proposed to train a CNN classifier to predict the realism of the composite image, and uses a gradient-based method to optimize the realism of the composite image. Recent works started to use end-to-end networks that output the harmonized composite image given the unharmonized image.
    \cite{tsai2017deep} is the first to use a CNN to perform the end-to-end transformation, where an encoder-decoder structure is adopted to capture the context and semantic information.
    \cite{cun2020improving} enhanced the neural network with a spatial-separated attention module that learns the features of specific regions in spatial space.
    \cite{guo2021image} adopted Vision Transformers which leverages its powerful ability to model long-range context dependencies.
    Different from all existing methods, we devote solving image harmonization with the diffusion model.
    
  
  \subsection{Diffusion Models}
    Diffusion models are a family of generative models that can produce realistic images from random noise, they have been shown to achieve state-of-the-art performance in image synthesis tasks\cite{dhariwal2021diffusion}.
    Ho et al.\cite{ho2020denoising} proposed denoising diffusion probabilistic models (DDPMs) based on a Markovian diffusion process that gradually adds noise to an image until it becomes pure noise, and then a deep neural network is trained as the noise estimator to reverses the process to generate a new image from the noise. 
    Song et al.\cite{song2020denoising} propose to use a non-Markovian diffusion process to train and sample from DDIMs, which is faster and more flexible than the Markovian diffusion process used in DDPMs.
    Due to the powerful ability to generate realistic images, diffusion models have been applied by researchers to a wide range of image synthesis tasks.
    \cite{rombach2022high} applied latent diffusion models (LDMs) to the text-to-image generation task, which are diffusion models applied in the latent space of pre-trained auto-encoders.
    Palette\cite{saharia2022palette} builds multi-task image-to-image diffusion models for end-to-end colorization, inpainting, uncropping, and JPEG restoration.
    SR3\cite{saharia2021image} presents a method for image super-resolution via repeated refinement.
    RePaint\cite{lugmayr2022repaint} uses a pre-trained unconditional DDPM as the generative prior and conditions the generation process by sampling the unmasked regions using the given image information.
    Blended Latent Diffusion (BLD)\cite{avrahami2022blended} uses LDMs for image editing, it allows users to control the generation process by specifying semantic attributes and blending them with the latent codes of existing images. 
    Similar to \cite{rombach2022high, lugmayr2022repaint} that the diffusion process is on the latent space, and image harmonization is similar to image-to-image translation.

\section{Proposed Method}
  In this section, we first discuss the proposed mechanism to force appearance consistency throughout the diffusion process of image harmonization.
  Then, we introduce a one-time color transfer method that does not change the diffusion process to ensure appearance consistency after the diffusion process.

  \subsection{Appearance Consistency Discriminator}
    \label{sec:acd}
    The appearance consistency discriminator evaluates how much two colored images are similar to each other in terms of appearance. 
    By converting the colored image to a grayscale image, we get to know the brightness information from the value of each pixel in the grayscale image, which can be seen as a representation of the appearance. For simplicity, we derive the grayscale image by averaging the RGB channels from the colored RGB image
    $$
    Y = C(X) = (X_R+X_G+X_B)/3,
    $$
    where $Y$ is the illuminance of the grayscale image, and $C$ is the function that converts images to grayscale.
    
    Our non parameter appearance consistency discriminator can be defined as
    $$
    D(X_1, X_2) = (C(X_1) - C(X_2))^2.
    $$ 
    The discriminator can capture only the difference in appearance between two images, which can guide the diffusion process to ensure appearance consistency and leave space for the noise estimator to adjust the color.
    
  \subsection{Classifier-guided LDM}
  \label{sec:guided}
    We derive the classifier-guided LDM from the classifier-guided DDIM algorithm \cite{dhariwal2021diffusion}. Different from the original algorithm, the gradients from the appearance consistency discriminator are passed through the discriminator.
    
    One of the challenges of using the classifier guidance in LDM is that the classifier should also work for noisy input. Since our method in \Cref{sec:acd} is nonparametric that we cannot further train it on the noisy data. Thus, we propose a method to enhance the capability of the discriminator on noisy inputs. Instead of calculating the guidance gradients between the reconstructed noisy image $x_t$ and one noisy guidance image $y_t$, we add noise to the guidance image multiple times to get multiple noisy guidance images $y_t^0, y_t^1, \dots, y_t^n$, and then calculate the gradients using all the images
    $$
        G(x_t, y) = \sum_{i=1}^n D(x_t, y_t^i), y_t^i \gets \mathcal{N}(\sqrt{\bar{\alpha}_t}x_0, (1-\bar{\alpha}_t)\textbf{I}).
    $$
    This avoids our reconstructed image guided by the random noise, instead focusing only on the useful information in the noisy guidance image, which remains the same in all possible noisy guidance images at the same timestep.
    \begin{algorithm}
  \caption{Classifier Guided LDM Sampling given a conditional diffusion model $\epsilon(h_t, c)$, classifier $p_\phi(y|x_t)$, encoder $\mathcal{E}(x)$, decoder $\mathcal{D}(h)$, and gradient scale s.}\label{alg:cap}
  \begin{algorithmic}
  \Require condition image $c$, gradient scale $s$
  \State $x_T \gets$ sample from $\mathcal{N}(0, \textbf{I})$
  \ForAll{$t$ from T to 0}
      \State $h_t\gets \mathcal{E}(x_t)$
      \ForAll{$i$ from 1 to n}
          \State $y_t^i \gets \mathcal{N}(\sqrt{\bar{\alpha}_t}x_0, (1-\bar{\alpha}_t)\textbf{I})$
      \EndFor
      \State $\hat{\epsilon} \gets \epsilon_\theta(h_t) - s\sqrt{1-\bar{\alpha}_t} \nabla_{x_t} \sum_{i=1}^nD(\mathcal{D}(h_t), y_t^i)$ 
      \State $h_{t-1} \gets \sqrt{\bar{\alpha}_{t-1}}(\frac{h_t - \sqrt{1-\bar{\alpha}_t}\hat{\epsilon}}{ \sqrt{\bar{\alpha}_t}}) 
      + \sqrt{1-\bar{\alpha}_{t-1}}\hat{\epsilon}$
      \State $x_t\gets \mathcal{D}(h_t)$
  \EndFor  
  \Ensure $x_0$
  \end{algorithmic}
\end{algorithm}

  \subsection{Color Transfer}   
    \label{sec:transfer}
    To only transfer the color information from one image to another, we need to separate the information contained in one image into appearance information and color information.

    Our method first converts the generated image from RGB space to HSV/HSL space. Since in the HSV/HSL space, there is a channel that represents the lightness information, which keeps the appearance of the image, we replace that channel with the same channel in the unharmonized composite image. Then we perform the linear transformation on the replaced channel based on the lightness channel of the background image to ensure the background and foreground have consistent brightness.

\section{Experiments}
  In this section, we demonstrate the harmonization capacity of our method, we evaluate our method with several experimental settings following the most representative latent diffusion model, stable diffusion \cite{Rombach_2022_CVPR}, and fine-tuned on the ControlNet \cite{zhang2023adding} framework. 
  
  \paragraph{Datasets}
    We evaluate our method on the iHarmony4 dataset, it is a comprehensive collection of synthesized composite images specifically designed for Image Harmonization research. It comprises four sub-datasets: HCOCO, HAdobe5k, HFlickr, and Hday2night. Each sub-dataset presents a distinct set of challenges and characteristics. The training set comprises a total of 65,742 samples, while the test set contains 7,404 samples. The synthesized composite images are generated using color transfer methods to transfer color information from reference images to real images. Four representative methods, selected from different categories based on parametric/non-parametric and correlated/decorrelated color space, were used.

  \paragraph{Models}
   For DDPM, we use the same U-Net model in \cite{dhariwal2021diffusion} as the noise predictor. For LDM, we follow the network architecture in ControlNet \cite{Rombach_2022_CVPR}, we added four stable diffusion encoder blocks and a middle block to the original network with skip connections, in which each block has the same shape as the original model. During the training, the weights of all the layers from the original stable diffusion model are frozen, and we only train the additional encoder and middle blocks.
   We use Adam optimizer with a learning rate 1$e$-5, and use a batch size of 4. We use Mean-Squared Errors (MSE) and PSNR scores on RGB channels as the evaluation metric. We report the average of MSE and PSNR over the test set. We resize the input images as 256 × 256 during both training and testing. MSE and PSNR are also calculated based on 256 × 256 images.

\section{Results}
  \subsection{Comparing with existing methods}
\begin{figure*}
 Mask   \qquad \qquad\qquad Ground Truth \qquad \quad Composite Image \qquad \quad \qquad Ours
  \centering
  \includegraphics[width=0.8\linewidth]{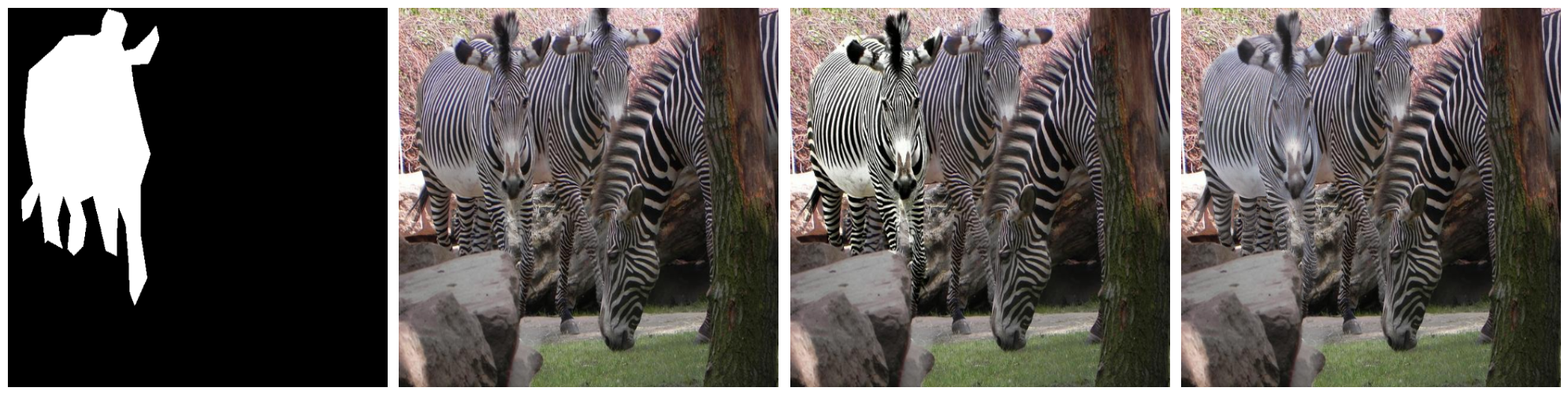}
  \includegraphics[width=0.8\linewidth]{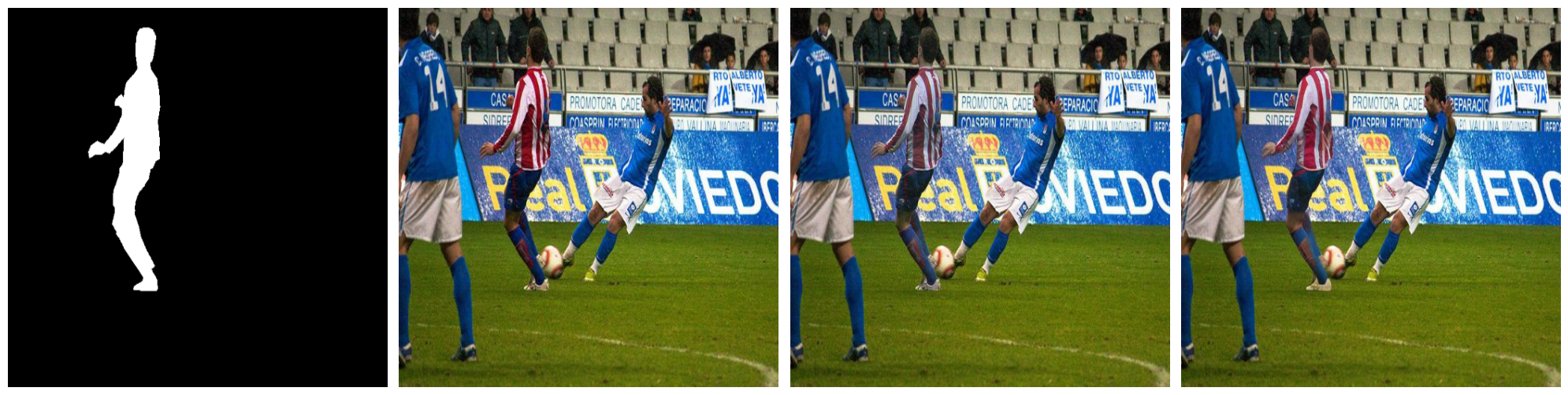}
  \includegraphics[width=0.8\linewidth]{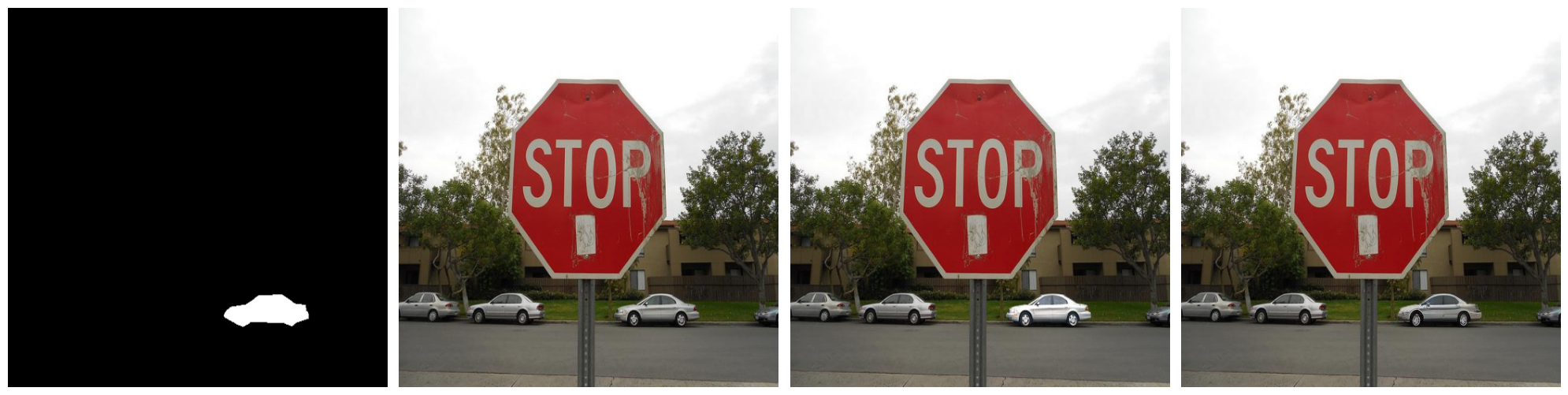}
  \includegraphics[width=0.8\linewidth]{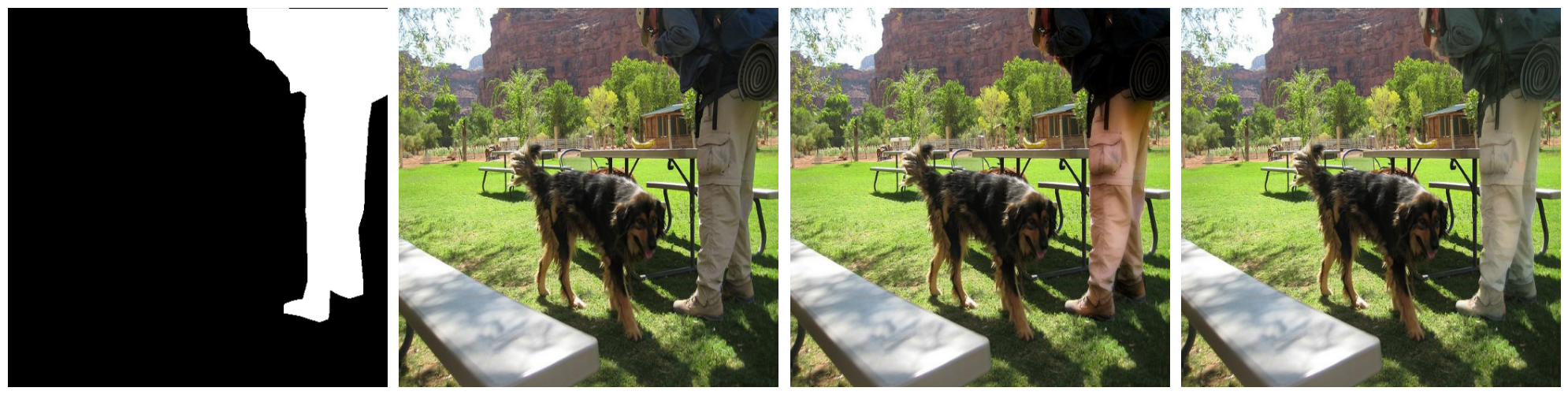}
  \includegraphics[width=0.8\linewidth]{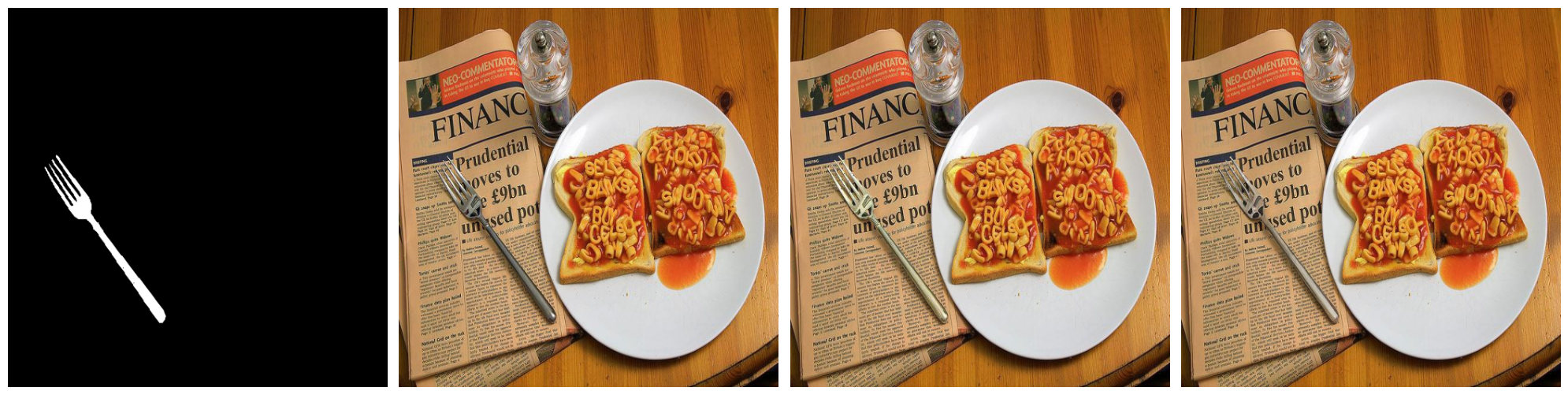}
  \includegraphics[width=0.8\linewidth]{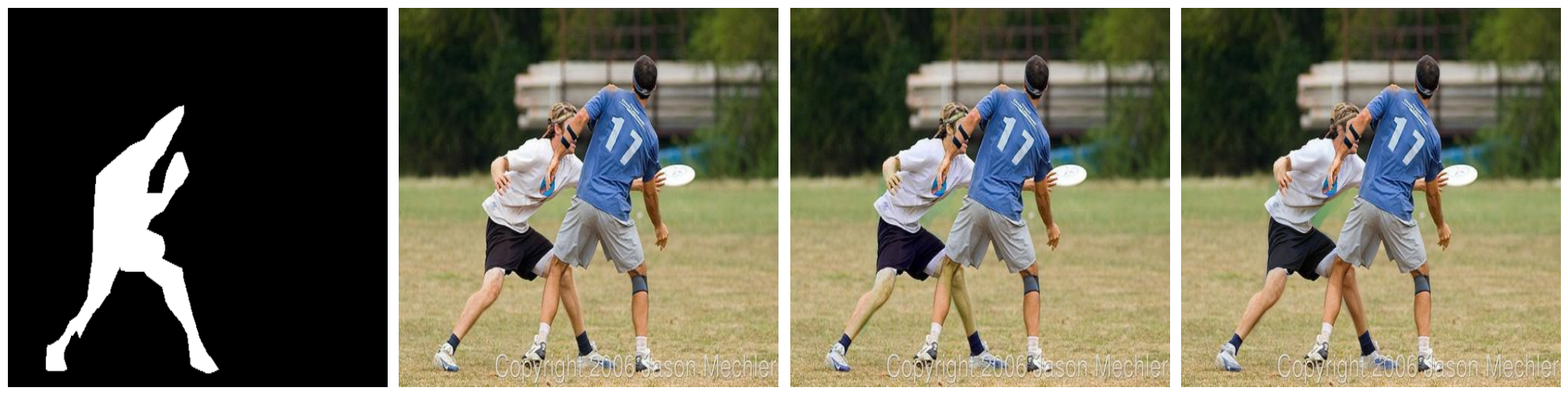}
  \caption{\textbf{Image harmonization results on HCOCO dataset.} From left to right, it is the mask, ground truth image, composite image and harmonized image (our method). 
  }
  \label{fig:test}
\end{figure*}

\begin{table}[ht]
\centering

\resizebox{\linewidth}{!}{%
  \begin{tabular}{@{}l|c|c|ccc|c@{}}
  \toprule
  Dataset     & Metric & Composite & DIH    & S$^2$AM & DoveNet & Ours  \\ \hline
  HCOCO       & PSNR   & 33.70     & 33.59  & 35.09   & 35.83   & 34.33 \\
              & MSE    & 70.39     & 56.17  & 35.65   & 34.26   & 59.55 \\ \hline
  HAdobe5k    & PSNR   & 28.31     & 32.36  & 34.23   & 35.13   & 33.18   \\
              & MSE    & 345.54    & 94.89  & 53.93   & 56.86   & 161.36  \\ \hline
  HFlickr     & PSNR   & 28.43     & 29.08  & 30.53   & 30.75   & 29.21   \\
              & MSE    & 264.35    & 168.35 & 123.36  & 125.85  & 224.05  \\ \hline
  Hday2night  & PSNR   & 34.36     & 33.59  & 34.48   & 34.87   & 34.08   \\
              & MSE    & 109.65    & 86.25  & 54.39   & 57.17   & 122.41  \\ \hline
  All         & PSNR   & 31.78     & 32.73  & 34.32   & 35.04   & 32.70   \\
              & MSE    & 172.47    & 80.55  & 51.13   & 51.51   & 141.84 \\ \hline
\end{tabular}
}
\caption{Quantitative comparison on iHarmony4.}
\label{tab:comparison}
\end{table}
    In Table~\ref{tab:comparison}, we provide the quantitative results of the MSE and PSNR scores of the harmonized images generated using the fine-tuned stable diffusion model guided by the appearance consistency discriminator in \Cref{sec:guided}, and post-process with color transferring method in \Cref{sec:transfer}. We compare the results with existing SOTA methods such as S$^2$AM \cite{cun2020improving} and DoveNet \cite{cong2020dovenet}. Visual results are presented in \Cref{fig:test}.
    
    \begin{figure*}
\quad Mask \quad \qquad  Composite Image \qquad \qquad   Ours \qquad \qquad \qquad CDTNet
  \centering
  \includegraphics[width=0.7\linewidth]{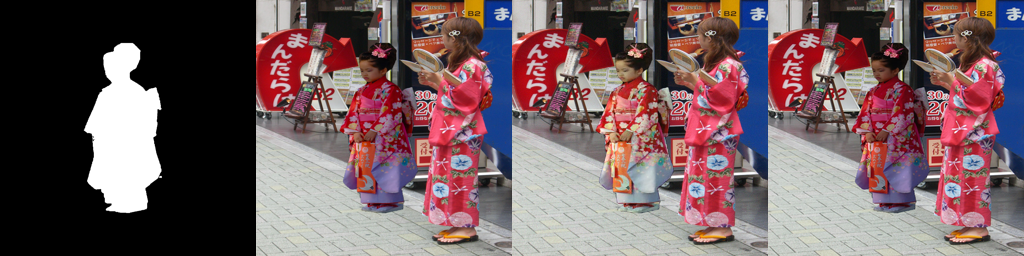}
  \includegraphics[width=0.7\linewidth]{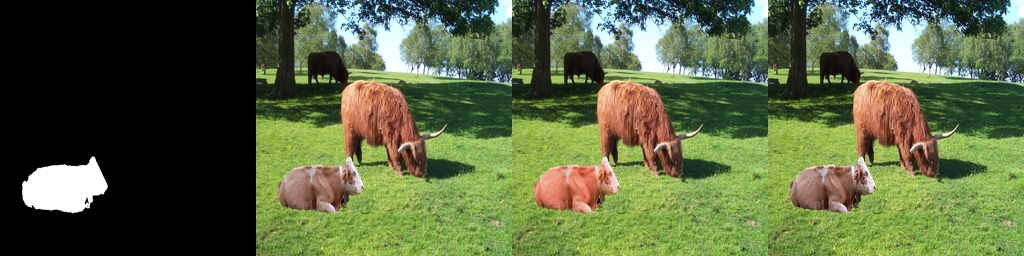}
  \includegraphics[width=0.7\linewidth]{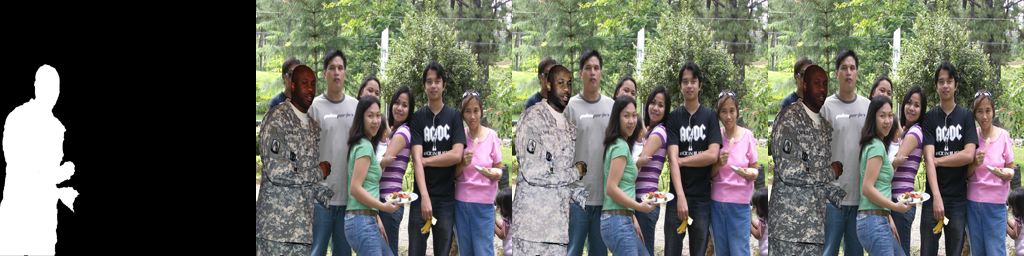}
  \includegraphics[width=0.7\linewidth]{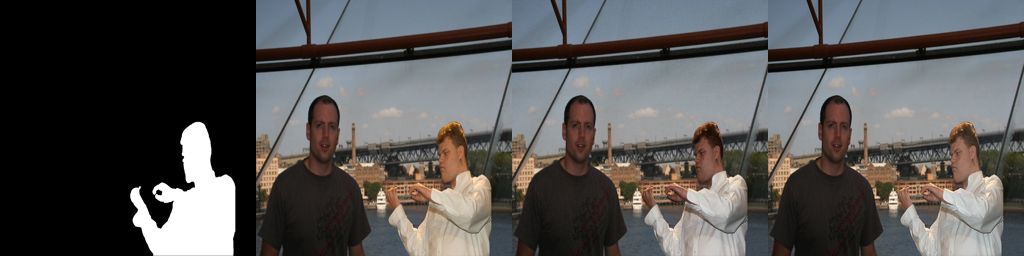}
  \includegraphics[width=0.7\linewidth]{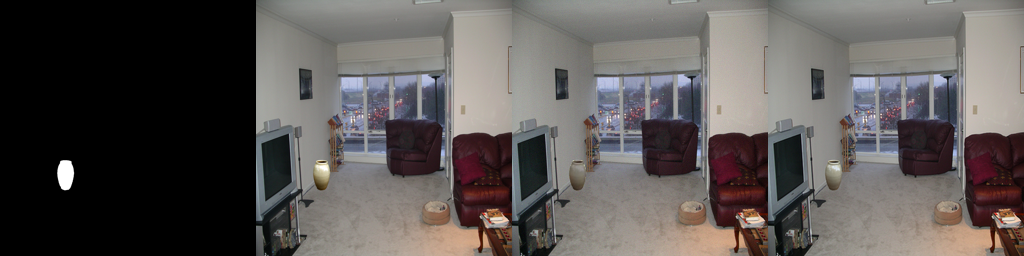}
  \includegraphics[width=0.7\linewidth]{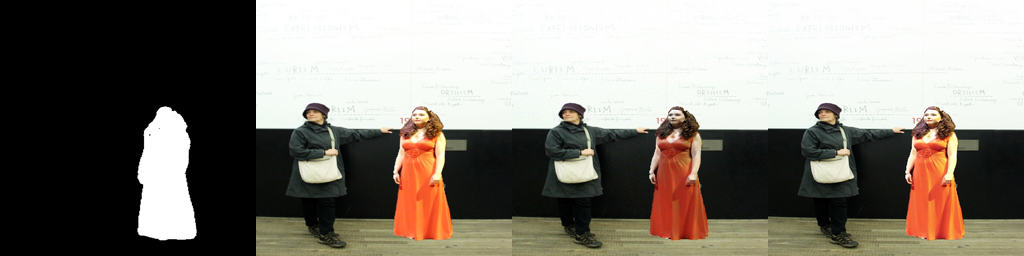}
  
  \caption{\textbf{Image harmonization results on real composite images.} From left to right, it is the mask, composite image and our result and CDTNet's result\cite{cong2022high}. Our results outperform the SOTA CDTNet in matching lightness, hat rendering, and object integration. We excel in accurately matching the lightness of persons and animals in the first four rows.
  Additionally, the last two images demonstrate improved object integration, with better alignment of the bottle and women to the background.}
  
  \label{fig:concat}
\end{figure*}

    \begin{figure*}[ht]
Mask \qquad \qquad Composite Image \qquad  \qquad Ground Truth \qquad \qquad \qquad Ours
  \centering
  \includegraphics[width=0.8\linewidth]{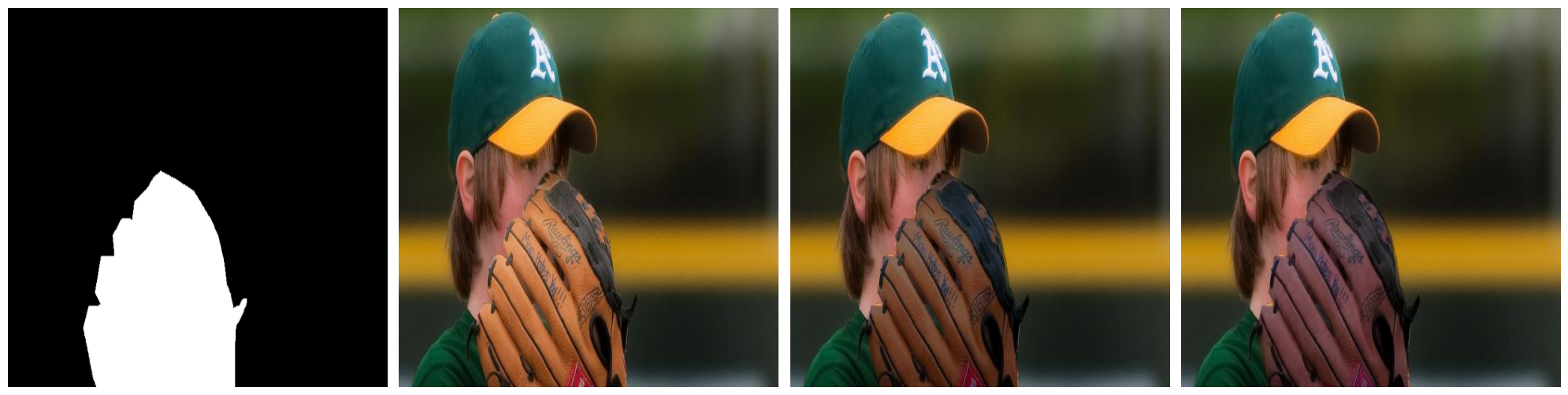}
  \includegraphics[width=0.8\linewidth]{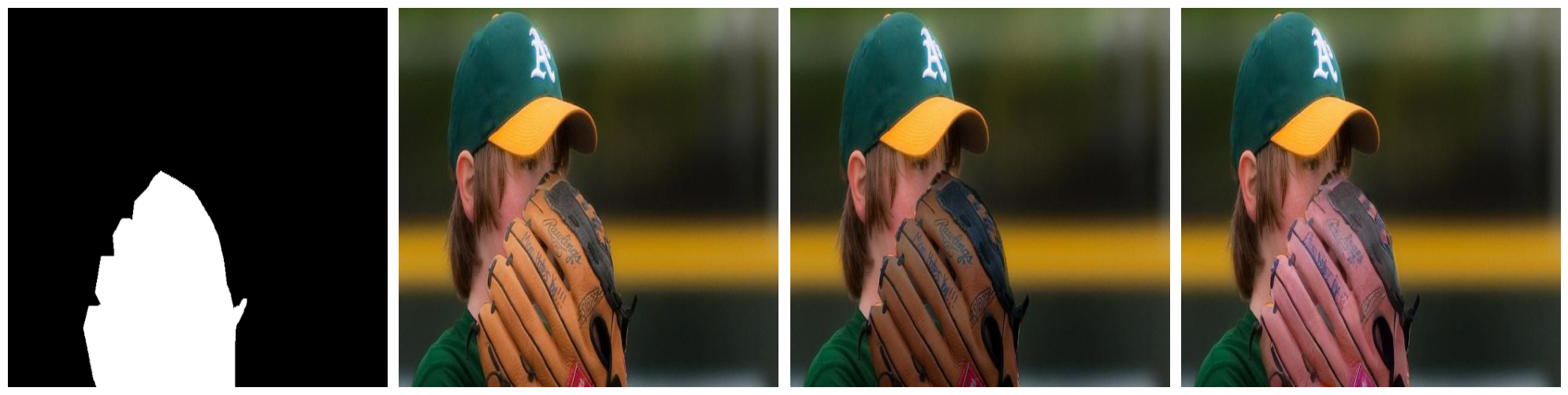}
  \caption{\textbf{Effectiveness of color transfer.} From left to right, it is the mask, ground truth image, composite image and harmonized image. The top images are the results with color transfer, while the bottom shows the results without color transfer. In the images above, the foreground image is the glove, without the color transfer, we found the characters on the glove is hard to read. By applying the color transfer method, the characters appear to be clear again.} 
  \label{fig:transfer}
\end{figure*}

    \begin{figure*}

\begin{flushleft} 
    \qquad Mask\quad\quad Composite Image \ CDTNet \qquad\qquad \qquad \qquad\qquad \qquad\qquad\quad Ours
    \vspace{-0.3cm}
\end{flushleft}

  \centering
  \includegraphics[width=1\linewidth]{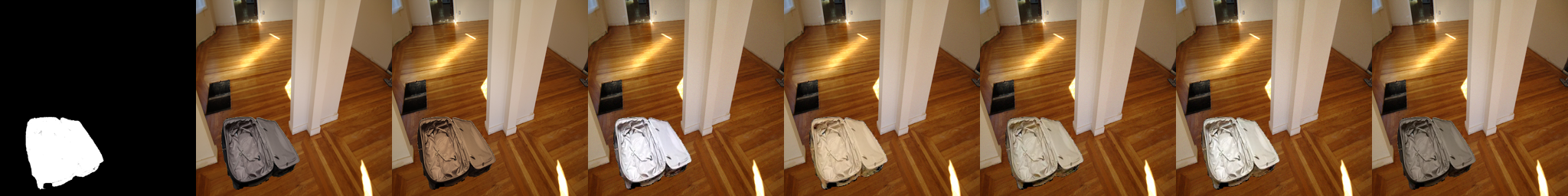}
  \includegraphics[width=1\linewidth]{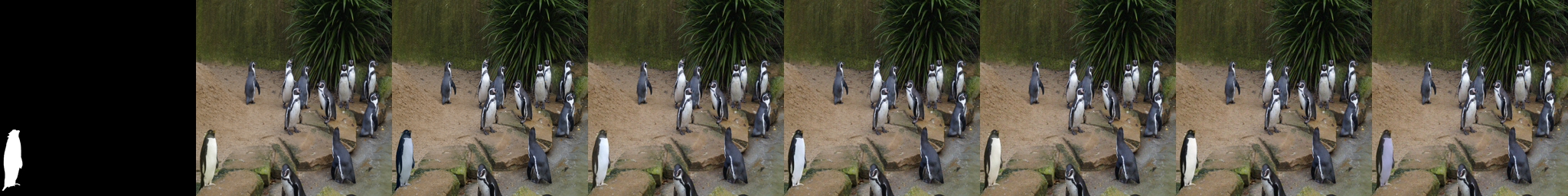}
  \includegraphics[width=1\linewidth]{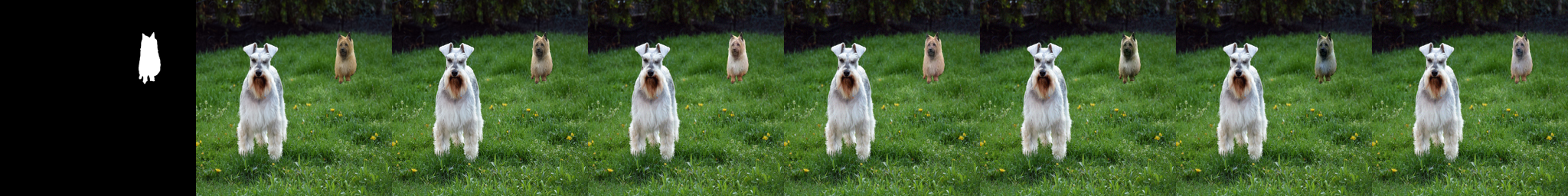}
  \includegraphics[width=1\linewidth]{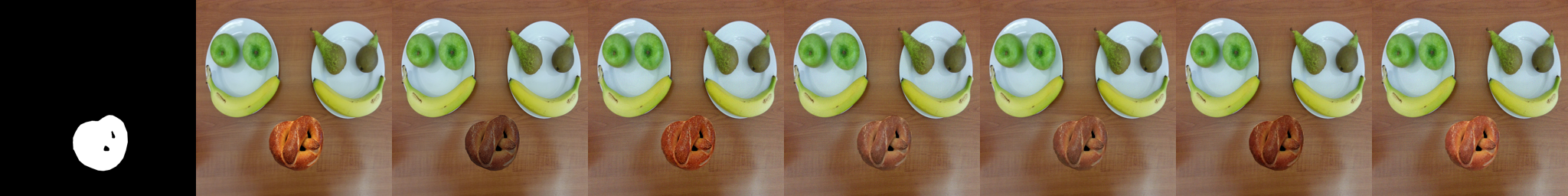}
  \includegraphics[width=1\linewidth]{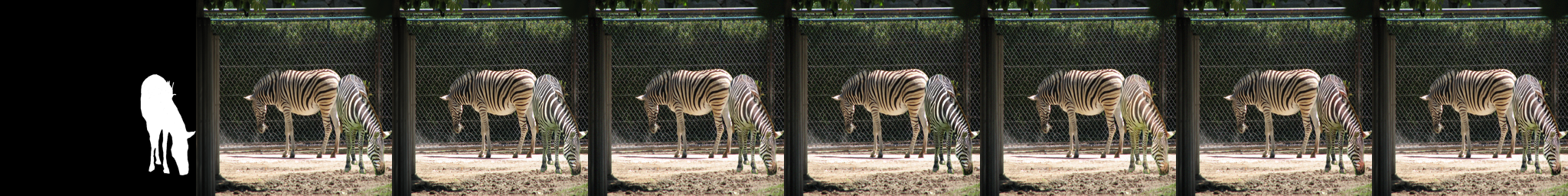}
  \includegraphics[width=1\linewidth]{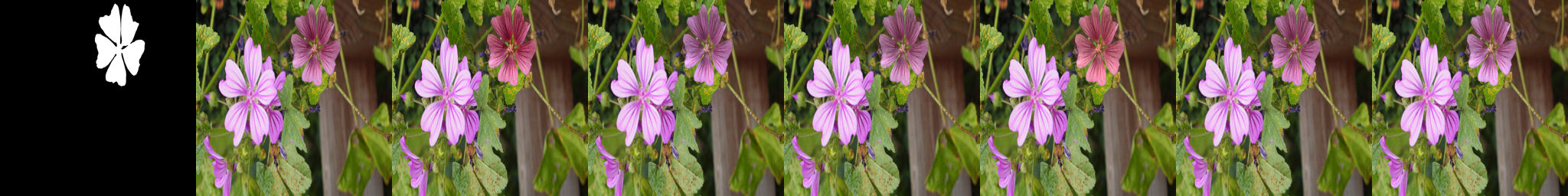}
  \includegraphics[width=1\linewidth]{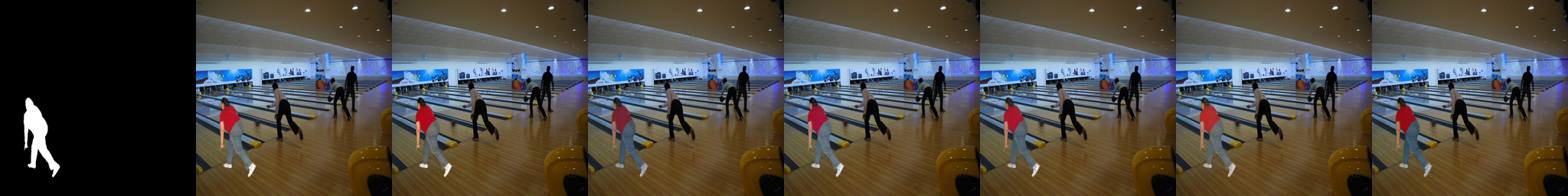}
  \label{fig:multi}
  
  \caption{\textbf{Examples of Multiple Output Results from the Diffusion Model.} We provide 5 results from our method for each composite image and compares them with CDTNet's result\cite{cong2022high}.}
\end{figure*}

    Unlike traditional deep learning approaches, the optimization objective of the diffusion model is not to directly minimize the difference between the generated image and the target image. Instead, it focuses on predicting the distribution of noise. As a result, the diffusion model may not outperform traditional metrics in terms of conventional evaluation measures. However, it still demonstrates the ability to generate high-quality harmonization results. 
    
    Furthermore, the iHarmony4 dataset presents a unique characteristic where composite images are generated by altering the colors of a portion of real images to create foreground images. While this provides a favorable learning target for the Diffusion Model, it also introduces a discrepancy between the dataset's input conditions and real-world usage scenarios, where foreground images typically originate from separate sources. To address this limitation and assess the performance of the models in realistic scenarios, we conducted additional experiments using synthesized composite images generated from the Open Image Dataset V6 and Flick Dataset. We compare our results with the current SOTA image harmonization model CDTNet\cite{cong2022high}. The results of these experiments are presented in \Cref{fig:concat}.

  \subsection{Effectiveness of color transfer}
    When applying LDMs to image editing tasks, one of the issues is the reconstructed image suffers from the reconstructing loss caused by the autoencoder. We show in \Cref{fig:transfer} that, our proposed method in \Cref{sec:transfer} can guide the stable diffusion model to maintain the consistency in appearance while only changing the color space of the composite image.

  \subsection{Multi-Output}
    Our Diffusion Model possesses an advantage over traditional end-to-end methods in that it exhibits inherent stochasticity. This enables us to generate multiple different results for the same input, providing users with a range of options to choose from. In \Cref{fig:multi}, we showcase several practical examples that utilize multiple output results from our model. By leveraging the stochastic nature of the Diffusion Model, we can offer users increased flexibility and control over the harmonization process. This capability allows for personalized and subjective adjustments, empowering users to select the output that best aligns with their preferences or specific requirements. The examples presented in Figure 7 highlight the diverse range of harmonized results that can be achieved through our model, demonstrating its effectiveness and potential for creative applications.

\section{Conclusion}
  In this paper, we have proposed a novel method for image harmonization based on diffusion models. Our method can effectively adjust the foreground image to match the background image in terms of illumination and color, resulting in realistic and harmonious composite images. We have conducted extensive experiments and ablation studies on synthesized image harmonization datasets and compared our method with existing methods. The results have shown that our method achieves state-of-the-art performance and outperforms the baselines by a large margin. Our method can be applied to various image editing tasks that require consistent lighting conditions. In the future, we plan to extend our method to handle real-world image harmonization scenarios, where the foreground and background images may have complex and diverse lighting conditions.

{\small
\bibliographystyle{ieee_fullname}
\bibliography{egbib}
}

\end{document}